\documentclass{article}

\usepackage[numbers]{natbib}
\usepackage[preprint]{neurips_2023}
\usepackage[utf8]{inputenc} 
\usepackage[T1]{fontenc}    
\usepackage{hyperref}       
\usepackage{url}            
\usepackage{booktabs}       
\usepackage{amsfonts}       
\usepackage{nicefrac}       
\usepackage{microtype}      
\usepackage{xcolor}         
\usepackage{graphicx}
\usepackage{subcaption}
\usepackage{verbatim}
\usepackage{mathtools} 
\usepackage{booktabs}
\usepackage{makecell}
\usepackage{caption}
\usepackage{amsmath}

\title{Evaluating the Capabilities of Multi-modal Reasoning Models with Synthetic Task Data}
\author{
  Nathan Vaska\thanks{corresponding author} \\
  MIT Lincoln Laboratory \\
  Lexington, MA 02421 \\
  \texttt{Nathan.Vaska@ll.mit.edu} \\
  \AND
  Victoria Helus\\
  MIT Lincoln Laboratory \\
  Lexington, MA 02421 \\
  \texttt{Victoria.Helus@ll.mit.edu} \\
}

\begin{document}

\maketitle
\setcounter{footnote}{0}

\begin{abstract}
The impressive advances and  applications of large language and joint language-and-visual understanding models has led to an increased need for methods of probing their potential reasoning capabilities. However, the difficulty of gather naturally-occurring data for complex multi-modal reasoning tasks bottlenecks the evaluation of AI methods on tasks which are not already covered by an academic dataset. In this work, we leverage recent advances in high resolution text-to-image generation to develop a framework for generating evaluation data for multi-modal reasoning tasks. We apply this framework to generate context-dependent anomaly data, creating a synthetic dataset on a challenging task which is not well covered by existing datasets. We benchmark the performance of a state-of-the-art visual question answering (VQA) model against data generated with this method, and demonstrate that while the task is tractable, the model performs significantly worse on the context-dependent anomaly detection task than on standard VQA tasks. 




\end{abstract}













\section{Introduction}
Language models, particularly large language models (LLMs) have demonstrated impressive performance on increasingly difficult reasoning tasks. At the same time, multi-modal models like BLIP-2 \citep{Li2023BLIP2BL} and GPT-4 \citep{OpenAI2023GPT4TR} which leverage LLMs have begun to show exceptional performance on tasks requiring both visual and textual reasoning such as the visual question answering task (VQA). This has led to increased interest in understanding the capability of these models to perform complex reasoning tasks involving multi-modal information \citep{Peng2023MMTransformer}. However, the process of gathering and labeling naturally occuring data to form datasets is both expensive and time consuming. This limits the tasks on which the reasoning capabilities of LLMs can be evaluated to existing academic datasets like VQAv2 and LSMDC \citep{Goyal2016MakingTV, Rohrbach2015ADF}. %

Context-dependent anomaly detection is one such unexplored reasoning task. A context-dependent anomaly occurs when an otherwise normal entity is significantly different from the context in which it is found. Unlike point anomalies, which can be identified by anomalous features of the entity (e.g. an incorrect coloration), reasoning about context-dependent anomalies requires a much broader understanding of both the normal characteristics of an entity and also the typical characteristics the entity with respect to its environment \citep{Pang2021DeepLF}. Detecting context-dependent anomalies is an open research task \citep{Pang2021DeepLF} and one at which LLM-based multi-modal models might excel given their powerful reasoning capabilities. However, to our knowledge no attempt has been made to benchmark multi-modal LLM models on context-dependent anomaly detection. This is likely due to the difficulty of finding data; as noted in \cite{Bozcan2021ContextDependentAD}, data generation for context-dependent anomalies is significantly more difficult than data generation for point anomalies due to their increased complexity. 

In parallel to the development of LLMs, the past few years have seen an increase in interest in the text-to-image generation task. This has resulted in the rise of diffusion models like DALL-E-2 \citep{Ramesh2022HierarchicalTI} and Stable Diffusion \citep{Rombach2021HighResolutionIS} which have set new performance benchmarks for quality and diversity for a variety of text-to-image generation tasks. Since these models can generate outputs for a diverse set of prompts, they offer a potential alternative for evaluating models against a specific reasoning tasks. However, this capability has not been widely explored.

In this work, we propose leveraging text-to-image generation capabilities to generate synthetic data for evaluating methods on data-starved multi-modal reasoning tasks. Using context-dependent anomaly detection as a representative task, we make the following contributions:

\begin{itemize}
    \item We propose a general framework for creating synthetic evaluation data for characterizing model performance on data-starved tasks.
    \item We apply our framework to generate an image context-dependent anomaly dataset that is 100 times larger than the most similar prior dataset. We use only publicly available data and models, a small amount of compute, and no human supervision.
    \item We frame image-based anomaly detection as a VQA task and identify a gap in the performance of a state-of-the-art, LLM based VQA model on the context-dependent anomaly detetion task using our synthetic data.
\end{itemize}

\section{Related work}

\textbf{Image generation}~~~The text-based image generation task takes a text description as an input and generates an image matching that description. Deep neural network-based image generation capabilities were initially popularized with the development of generative adversarial networks (GAN) \citep{goodfellow2020generative}, which could generate images from their training distribution. While GAN methods have been extended to the text-based generation task, more recently diffusion models like DALL-E-2 and Stable Diffusion have surpassed GAN methods on common benchmarks \citep{Rombach_2022_CVPR}. Diffusion models learn to reconstruct an image from a latent space representation using a learned reverse diffusion process. They have also been applied to text-based inpainting, in which an object or region of an existing image is replaced with new visual information using a diffusion model conditioned on text \citep{Rombach_2022_CVPR}. 


Despite only developing relatively recently, there is growing body of work applying generative diffusion models to augment or replace training data. Efforts have generated synthetic training data with Stable Diffusion and trained ImageNet classifiers \citep{bansal2023leaving} and \citep{sariyildiz2023fake}, fine-tuned the Imagen diffusion model to produce better training data for ImageNet classifiers \cite{azizi2023synthetic}, and used the GLIDE\citep{Nichol2021GLIDETP} model to improve few/zero shot classification on CIFAR-100 \cite{he2022synthetic}. \cite{Trabucco2023EffectiveDA} is the closest to our approach as they use diffusion models to augment individual images; however, their focus is on improving few-shot detection rather than generating evaluation data. To our knowledge, no prior work has explored the use of diffusion models for generating evaluation data on novel tasks.

\textbf{Multi-modal models}~~~There are a wide variety of multi-modal models; in this work we will consider both joint embedding models and visual question answering (VQA) models. In the joint visual-language embedding task, the goal is to generate similar embeddings for images and text that share semantic information. The CLIP architecture \citep{Radford2021LearningTV} approached this task using a constrastive loss; other methods for joint embedding utilize more complex methods \cite{sun2023eva}. Joint embedding models have been incorporated into a variety of other tasks, including enabling zero shot detection in object detectors \citep{Carion2020EndtoEndOD}, generating semantic embeddings for text-based image generation \citep{Rombach2021HighResolutionIS}, and retrieving knowledge for VQA tasks \citep{Gui2021KATAK}. In the visual questioning answering (VQA) task, the objective is to correctly answer a question based on visual information. While initially formulated as a classification style task in which correct answers were predicted from a fixed set, current VQA benchmarks \citep{Agrawal2015VQAVQ} are based on evaluating the similarity of free-form text to the set of correct answers. Open knowledge VQA tasks, first introduced in \cite{Marino2019OKVQAAV}, is another variant of the task in which the question cannot be correctly answered solely based on the image; external knowledge must be combined with visual information to arrive at the correct answer. However, both standard and open knowledge VQA datasets do not have good coverage of questions relating to anomaly detection. For example, an inspection of the \textasciitilde14,000 training questions in the OK-VQA dataset yields only six questions that relate to anomaly detection.

\textbf{Anomaly detection}~~~Anomaly detection is the task of detecting observations that differ from the standard distribution of data and is relevant in many fields, including cybersecurity, finance, and healthcare. Point anomalies, or anomalies where an individual instance of an entity differs from normal instances, are the most commonly studied type of anomaly. Deep learning methods have demonstrated performance on detecting a variety of point anomalies \citep{Pang2021DeepLF}. However, there are several other types of anomalies which have been less well studied. One such type of anomaly is a contextual anomaly, which is a normal instance of an entity that is anomalous with respect to its current context \citep{Pang2021DeepLF, Chandola2009AD}. Performing context-dependent anomaly detection usually involves trying to reduce the problem to a point anomaly detection problem, where multiple models may be conditioned on specific contexts, which may not be feasible if the class space is large \citep{Bozcan2021ContextDependentAD, Chandola2009AD}. Aside from \cite{Bozcan2021ContextDependentAD}, which evaluated a method for detecting context-dependent anomalies in images on a small, aerial photo centric dataset and \cite{Vaska2022ContextDependentAD} which used knowledge graphs to detect anomalies in data derived from images but did not utilize visual information, there is little work on detecting context-dependent anomalies in images.

\section{Method}

In order to generate datasets for complex reasoning tasks on visual data, the outputs of the data generation process must be consistent with respect to the process inputs. This property allows the label of any generated data to be derived from the pipeline's input, which is critical for removing the need for human labeling. Additionally, the outputs of the pipeline should be close to natural images as possible so that the dataset does not introduce an additional distribution shift that could affect model performance. 

To meet these requirements, we propose a data generation framework in which a targeted region of a natural image is replaced with an image generated using a text description that corresponds to the desired task. Since the majority of the image is naturally occurring data, as long as the edited region correctly corresponds to the text description, this approach achieves both properties. One constraint on this approach is that an appropriate input image must be available in sufficient quantities; however this constraint does not pose a severe restriction on this method as there are a plethora of natural image datasets at different levels of labeling available. 

To ground our framework in an specific task, we will focus on generating data for a context-dependent anomaly detection task. Under this task definition, our framework's goal is to replace a region of a image with visual information that is out of place compared to the original image. The following sections describe the specific anomaly detection task and the details of our approach.

\subsection{Synthetic anomaly generation}

Let us define our class of contextual anomalies as objects which do not belong in the indoor scene type in which they are found (e.g. a pizza is anomalous in a bathroom but not a kitchen). More specifically, let $O$ be a set of potential objects and $S$ a set of possible scenes. Consider an image which depicts a scene $s_i \in S$ and a number of objects $O_i = [o_1, ..., o_n]$ where $O_i \subset O$. The image also contains a single anomalous object $o_a \in O$ that is anomalous in the context of $s_i$ and $O_i$. The context-dependent anomaly detection task can be defined as correctly identifying $o_a$ from $o_a \cup O_i$. Note that under this task definition every image is guaranteed to contain one and only one anomaly. 


\subsubsection{Injecting targeted anomalous visual information}
The first step of generating a synthetic context-dependent anomaly dataset is to convert non-anomalous images to anomalous images such that the resulting image is labeled with the injected anomaly. Let $q$ be a non-anomalous image with an associated scene type $s$ and a masked region $m$. For each $s \in S$, $\exists O_s \subset O$, where $O_s$ is a set of objects that are anomalous with respect to that scene. Select an anomalous object $o_a \in O_s$. Let $g$ be a text-to-image inpainting model which takes a target object, an image, and a masked region as input. Its output is the anomalous object's corresponding visual representation, which we denote $v_a$: 


\[
g(o_a, q, m) = v_a
\]

Within $q$, replace $m$ with $v_a$ to create a new candidate image $p$, in which $o_a$ is a known to exist as a context-dependent anomaly. 

\subsubsection{Ensuring consistency of data generation with filtering}
Text-to-image generation models may generate outputs corresponding to a different class or introduce additional artifacts which significantly distort the generated visual information, generating an undesirable distribution shift. To improve the quality of samples generated by our pipeline, we utilize a joint visual-language embedding model to filter out generations which differ significantly from the intended target.  

Let $P$ be an image dataset of candidate anomalous images. We consider each image as a potential addition to $P_A$, our desired, filtered dataset that is comprised of images that are more realistic representations of our anomalies. Given a joint visual-language embedding model $f$, we compute a set $Z$ of similarity scores between the the visual information of the image's anomalous object $v_a$ and every object label $o_i \in O$, including $o_a$:

\[
\begin{aligned}
\!
f(v_a, o_i) = z_i  \\
f(v_a, o_a) = z_a  \\
Z = \{z_a \cup \bigcup\limits_{i=1}^{|O|} z_i\}
\end{aligned}
\]


Then each element $p \in P$ is an input image which we represent as $p = (o_a, v_a, Z)$ to indicate included information about the image's anomalous object, visual representation of said object, and its associated similarity scores. We use these similarity scores to determine whether $p$ is likely to be visually realistic or aligned with the target object.

If the similarity score $z_a$ is in the top-$k$ similarities, then we include $p$ in $P_A$. Additionally, consider a set of objects $O_t \subset O$, where a high similarity between $v_a$ and an object in $O_t$ indicates the presence of an artifact. For example, in our experience, human body parts (i.e., arms, legs) would appear in some generated images. To prevent inadvertently including images that contain artifacts, we would want to make sure that the scores corresponding to these objects in $O_t$ are low. If the similarity scores $f(v_a, o_t) = z_t \forall o_t \in O_t$ are not in the top $k$, then we can include $p$ in $P_A$. This process is repeated for every image in $P$:

\[
    \text{top-}k(z) =
    \begin{cases} 
      1 & \text{if $z$ is in the top-$k$ elements of $Z$}\\
      0 & \text{otherwise}
    \end{cases}
\]

\[
P_A = \{p \in P : \text{top-}k(z_a) = 1 \: and \: \text{top-}k(z_t) = 0 \,\forall o_t \in O_t \} 
\]

\section{Experiments}

Figure \ref{fig:gen_process} is a high-level illustration of our data generation pipeline. We first apply our pipeline to generate an example dataset for the indoor objects context-dependent anomaly detection task. We demonstrate the value of the synthetically generated data by comparing the performance of current VQA models to a simple, similarity-based anomaly detection method on our task, highlighting gaps in the reasoning capability of the VQA models. We provide details on the dataset generation and evaluation methods below.

\subsection{Dataset generation}

\begin{figure}
    \centering
    \includegraphics[width=\textwidth]{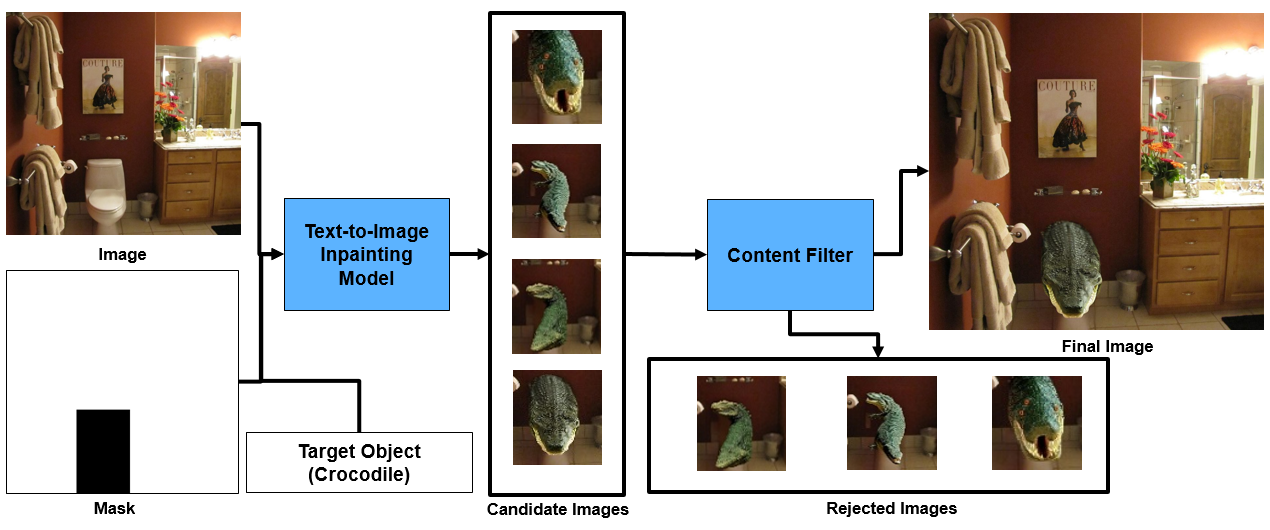}
    \caption{Anomaly generation pipeline. Given that an image of a bathroom has been selected and a large object has been masked, crocodile is selected as the anomalous object. An off the shelf inpainting model is used to create several candidate replacement images, and embedding based content filters are used to remove examples which are too different from the target object.}
    \label{fig:gen_process}
\end{figure}

We used the set of objects from the Open Images object detection dataset \cite{kuznetsova2020open} as our overall set of object classes. We identified eight different scene types which presented different indoor contexts, selected 50 objects from our overall object set to serve as anomaly targets, and characterized whether each object would be anomalous in each scene type. We also categorized each object's size as small (apple-sized), medium (microwave-sized), or large (appliance-sized). We chose to limit the number of anomalous object classes to allow more oversight into the generated data; however, our method can easily be extended to additional anomalous object since the only human input required is a size classification and a small number of yes/no decisions on which scene types an object is anomalous in. The full list of anomalies and scene types can be found in Appendix. We then filtered the Visual Genome (VG) dataset \cite{krishna2017visual} for images matching the selected scene types and chose an unbalanced selection of 80 base images. We manually selected images to ensure that a large number of objects present were present in each image, since cluttered scenes have large amounts of task relevant context. Images could have also been selected automatically based on dataset criterion like the number of object detections; however, we did not explore automatic image selection in this work. For each image, we applied 3-4 rectangular masks around objects in the scene and coded each mask as small, medium, or large using the same criterion as the objects. 

For each image and mask pair, we selected the set of objects considered anomalous for the image's scene type. We then removed any objects which did not have the same size label as the mask, as we found that the inpainting model we used, Stable Diffusion inpainting \citep{Rombach2021HighResolutionIS}, often distorted the size of the object to match the size of the mask even if this greatly changed the object proportions relative to the scene. We also found that object replacement was more consistent when the object was the focal point of the scene, so we cropped images to a window around the masked region. As shown on the left side of \ref{fig:gen_process}, we then ran each cropped image, mask, and object triplet through our chosen text-based inpainting model to generate 10 candidate replacement images for the masked area. Candidate images were filtered both for lack of object consistency and for the presence of human attributes (arm, leg, etc), which often indicated a distorted generation, which is shown on the right side of \ref{fig:gen_process}. Filtering was done using a full object description instead of just the object name (e.g. "apple: a red fruit ..." instead of apple), as we found that this improved filtering quality, and with $k=5$. Any candidate images that passed through the filtering were injected into the original image in place of the mask to generate the final anomalous images. 

\subsection{VQA-based anomaly detection}

VQA methods take an image and a question as input, and return an answer to the question based on the image content as an output. To frame our image-based context-dependent anomaly detection as a VQA task, no major modifications are needed. It is sufficient to write the query posed to the VQA model such that, given an image containing a single anomalous object, the the correct answer to the query will correspond to the anomalous object.

\subsection{VQA evaluation methods}


Calculating the accuracy of the VQA model on the anomaly detection task is non-trivial since the model returns free form text. To correctly calculate accuracy, the text outputs must be accurately mapped to the object class labels. Previous VQA datasets have handled this by giving weighted credit based on the number of human participants who gave each answer. Since our pipeline only provides exact class labels, this method is not directly applicable; we instead utilize three different evaluation tactics. First, we do a VQA-like direct word matching, giving the model credit for a response if the response contains the true answer as a discrete word. As this method penalizes responses which use synonyms (i.e. llama instead of alpaca) we also use an embedding language model to perform zero shot classification on the response, mapping it to the classes in the Open Images object detection dataset. This handles synonyms and related concepts better than the VQA-like method, but may incorrectly penalize the model for predicting a semantically close object. To address this issue, we also perform a zero shot prediction over 10 representative classes (i.e. tool, food). This metric is unlikely to penalize the model for predicting semantically similar classes, but does greatly simplify the anomaly prediction task. For the zero shot prediction methods we also generate and append a description of the category or model response when generating the embeddings; we found that this qualitatively improves the matching. 

\subsection{Similarity-based anomaly detection}
A valid concern when evaluation models using synthetic data is that the targeted task may be intractable if quality of the generated data is poor. For instance, if the goal is to identify that an apple is anomalous in a particular image but the apple looks more like a baseball, there is no viable way for an anomaly detection model to correctly identify the anomaly with the correct reasoning. In order to better understand whether the data generated by our pipeline is sufficiently high quality for the anomaly detection task to be tractable, we develop a simple feature-based method for detecting context-dependent anomalies in image data. 

At a high-level, the similarity-based anomaly detection method uses a region proposal network to split the image into regions corresponding to objects, applies at set of similarity functions to calculate scores representing how similar each region is to the a component of the image's context, and then combines the scores into an overall similarity score. The anomaly detection problem is then reduced to selecting the region with the lowest similarity score and predicting the class of the object corresponding to that region. This approach is similar to the method used in \cite{Vaska2022ContextDependentAD}, and full details on this method can be found in Appendix.

The similarity functions drive the reasoning capabilities of this approach. Intuitively, similarity functions can be definitely with visual features (a chair looks like a table) or semantic features (a chair is semantically similar to a table). Visual features can be extracted directly from image regions using an image embedding model. Calculating semantic features for a region requires associating outside knowledge with that region; we associate text-based semantic knowledge from a knowledge base with each region following the approach from \cite{Gui2021KATAK}. We then calculate the semantic features from the knowledge using a text embedding model. Given visual features and semantic features, similarity functions can be defined to compare regions to each other and the full image to calculate a score representing how similar each region is to the overall scene context. We define three sets of similarity functions for our evaluation: functions which use only semantic features (Knowledge), functions which use only visual features (Visual), and the set of all functions (All).

\begin{table}[!t]
\caption{Comparison of Generated Dataset to Existing Image Anomaly Datasets}
\centering
\begin{tabular}{lcccl}\toprule 
Dataset & Domain & \makecell{Anomalous \\ Samples}  & \makecell{Anomaly\\ Type} & \makecell{Generation \\ Method} \\\midrule
\makecell[l]{MVTec AD \citep{bergmann2019mvtec}} & Defect Detection  & 1258  & Point & Image Collection \\
\makecell[l]{UCI letter \citep{han2022adbench}}   & Handwriting       & 100   & Point & Data Subsampling \\
\makecell[l]{UCI opt.digits \citep{han2022adbench}} & Handwriting     & 150   & Point & Data Subsampling\\
\makecell[l]{UCI skin \citep{han2022adbench}} & Human Features        & 50859 & Point & Data Subsampling \\
\makecell[l]{AU-AIR Anomaly  \citep{Bozcan2021ContextDependentAD}}  & Traffic Surveillance & 120 & Contextual &  Manual Editing \\
 \bottomrule
Ours & Household Images & 10,527 & Contextual & \makecell[l]{Text Based \\ Inpainting} \\ 
\end{tabular}
\label{tab:dataset_comp}
\end{table}

\begin{figure}[!t]
    \centering
    \begin{subfigure}[b]{0.24\textwidth}
        \centering
        \includegraphics[width=.9\textwidth]{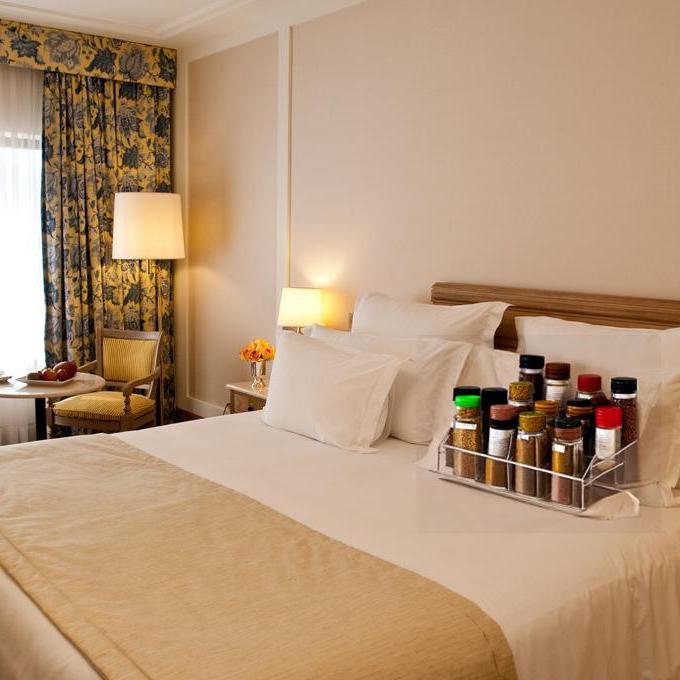}
        \caption{\makecell[l]{spice rack in a \\ hotel room}}
    \end{subfigure}
    \begin{subfigure}[b]{0.24\textwidth}
        \centering
        \includegraphics[width=.9\textwidth]{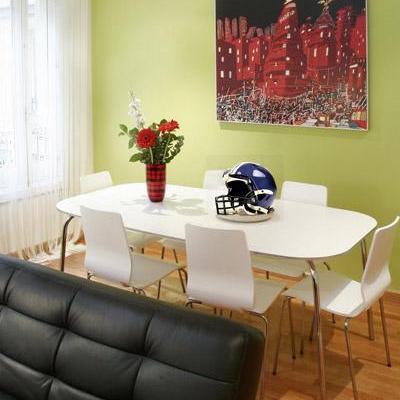}
        \caption{\makecell[l]{helmet in a \\dining room}}
    \end{subfigure}
    \begin{subfigure}[b]{0.24\textwidth}
        \centering
        \includegraphics[width=.9\textwidth]{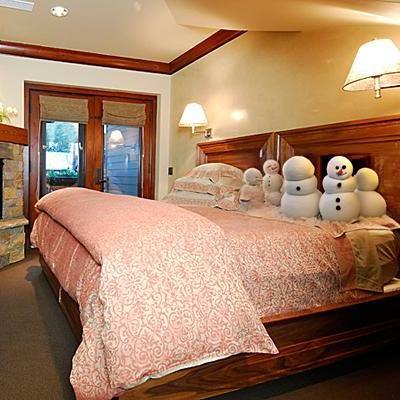}
        \caption{\makecell[l]{snowman in a \\bedroom}}
    \end{subfigure}
    \begin{subfigure}[b]{0.24\textwidth}
        \centering
        \includegraphics[width=.9\textwidth]{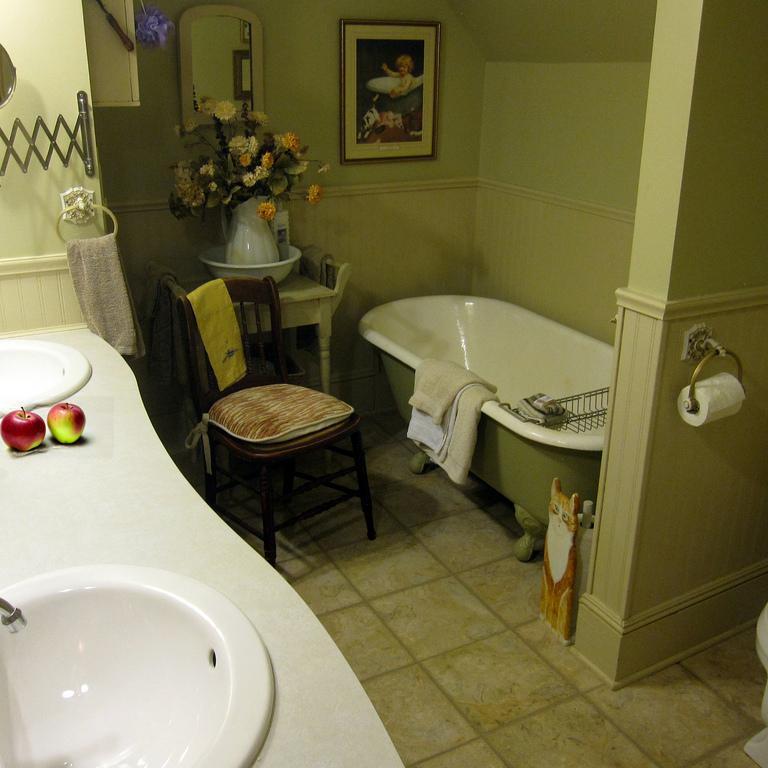}
        \caption{\makecell[l]{apple in a \\bathroom}}
    \end{subfigure}
    \\[\smallskipamount]
    \begin{subfigure}[b]{0.24\textwidth}
        \centering
        \includegraphics[width=.9\textwidth]{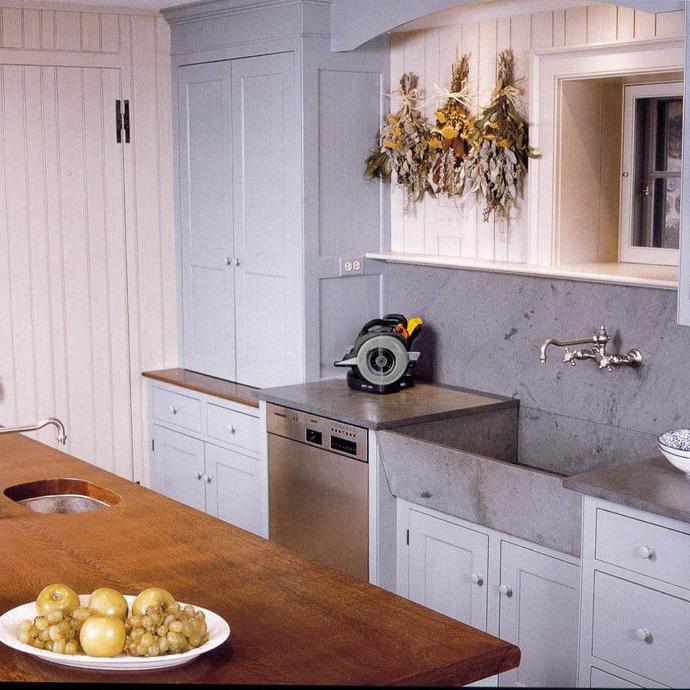}
        \caption{\makecell[l]{chainsaw in a \\kitchen}}
    \end{subfigure}
    \begin{subfigure}[b]{0.24\textwidth}
        \centering
        \includegraphics[width=.9\textwidth]{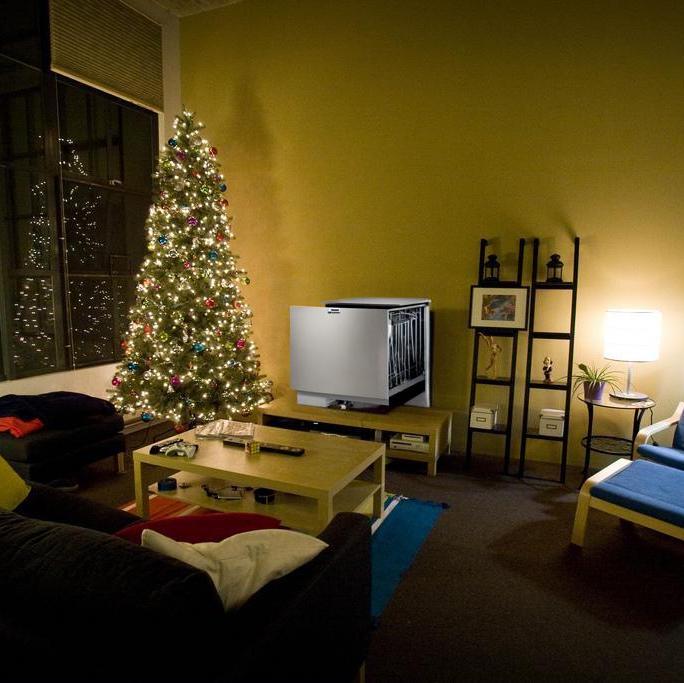}
        \caption{\makecell[l]{dishwasher in a \\living room}}
    \end{subfigure}
    \begin{subfigure}[b]{0.24\textwidth}
        \centering
        \includegraphics[width=.9\textwidth]{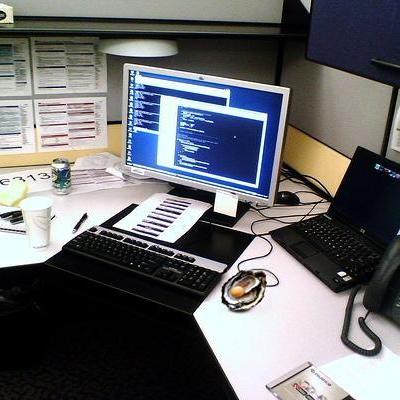}
        \caption{\makecell[l]{oyster in a \\cubicle}}
    \end{subfigure}
    \begin{subfigure}[b]{0.24\textwidth}
        \centering
        \includegraphics[width=.9\textwidth]{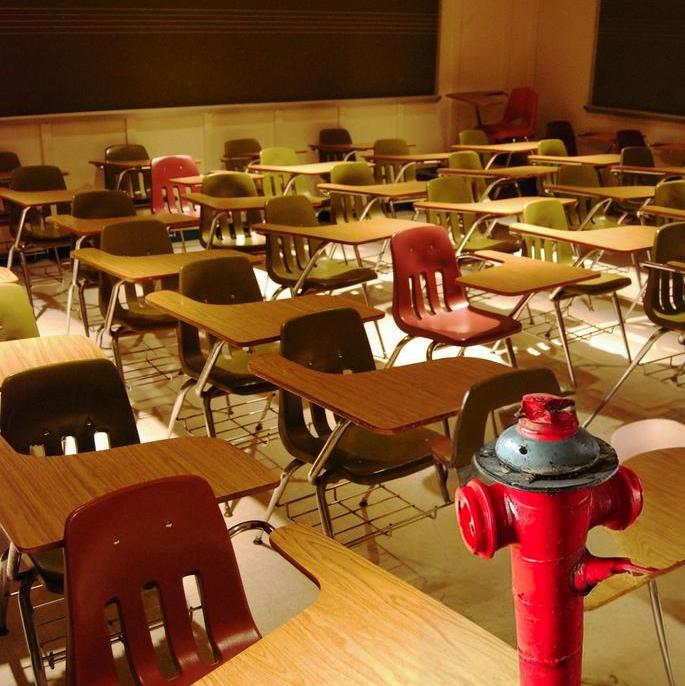}
        \caption{\makecell[l]{fire hydrant in a \\classroom}}
    \end{subfigure}
    \caption{Examples generated using pipeline. A single context-dependent anomaly has been injected into each image at its native resolution. For more samples, see Appendix.}
    \label{fig:examples}
\end{figure}

\subsection{Model details}

A variety of models are required for the similarity and VQA-based anomaly detection models. For the similarity-based method, a region proposal network and a joint-visual language embedding network are required. For our experiments, we utilized the DETR model's \citep{Carion2020EndtoEndOD} region proposal network and the CLIP model \cite{Radford2021LearningTV} for joint visual and language embedding in the similarity functions and evaluation metrics . Additionally, the knowledge-based similarity function and our evaluation method required object descriptions for all Open Images objects. We utilized the text generation capabilities of ChatGPT \citep{OpenAI2023GPT4TR} to generate these descriptions and manually verified that the descriptions were accurate. See the Appendix for additional details on this process. 

For the VQA-based anomaly detector, we chose to utilize the recent BLIP-2 model. BLIP-2 has demonstrated state-of-the-art performance on standard VQA benchmarks and, in contrast to many other competitive VQA models, has open implementations available. We utilized the FlanT5-XL and FlanT5-XXL BLIP-2 model variants that had not been fine-tuned on other VQA datasets to avoid any dataset-specific biases. For model prompting, we experimented with a variety of different prompt wordings, ultimately using the following question wording: \textit{"Question: A context-dependent anomaly is an object that is anomalous based only on the context in which it is found. What object is the context-dependent anomaly in this scene? Short answer: "}. Other hyperparameters for text generation were set to the default values provided in the BLIP-2 paper.  

\section{Results}

\subsection{Dataset generation Results}

\begin{table}[t]
\caption{Performance of VQA-based method on contextual anomaly detection}
\centering
\begin{tabular}{lcccccl}\toprule 

& VQA-like Accuracy & \multicolumn{2}{c}{\makecell{Matching Accuracy \\ with Descriptions}} & \multicolumn{2}{c}{\makecell{Broad Category Matching \\ Accuracy with Descriptions}}
\\\cmidrule(lr){2-2} \cmidrule(lr){3-4} \cmidrule(lr){5-6}
Model Variant        & Top 1  & Top 1     & Top 3    & Top 1  & Top 3 \\\midrule
BLIP-2 Flan T5 xl    &  10.54    &  11.61    & 11.94    &  21.63  & 46.57 \\
BLIP-2 Flan T5 xxl   &  12.31    &  13.41    & 13.92    &  22.13  & 48.42 \\
 \bottomrule
\end{tabular}
\label{tab:blip_acc}
\end{table}

Two A100 GPUs were used for data generation, and all data was generated during an \textasciitilde8 hour generation run. Of the 25,204 instances generated with the pipeline, 10,527 were accepted by the content filters resulting in an acceptance rate of 41.8\%. Figure \ref{fig:examples} shows example images altered using our data generation pipeline; qualitatively, the generated images align with the desired classes. Additional example generations, including instances that were rejected based on our content filters, are shown in the Appendix. Table \ref{tab:dataset_comp} compares the generated dataset to existing anomaly datasets that relate to image data; the generated dataset is about 100x the size of the only other image-based contextual anomaly dataset despite using a low amount of compute and no human supervision.

\begin{figure}[t]
    \centering
    \begin{subfigure}[b]{0.3\textwidth}
        \centering
        \includegraphics[width=.9\textwidth]{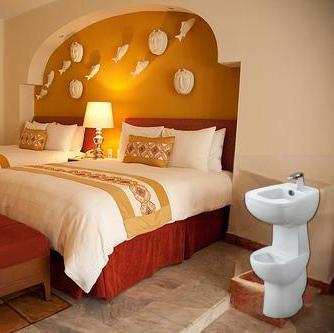}
        \caption{There are two beds in the room}
    \end{subfigure}
    \begin{subfigure}[b]{0.3\textwidth}
        \centering
        \includegraphics[width=.9\textwidth]{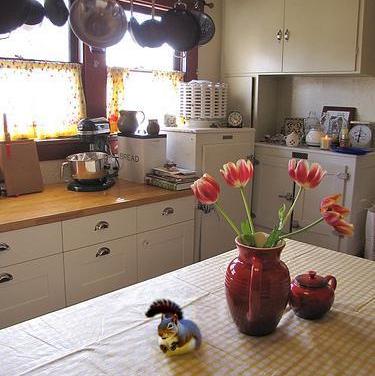}
        \caption{tulips in the kitchen}
    \end{subfigure}
    \begin{subfigure}[b]{0.3\textwidth}
    \centering
        \includegraphics[width=.9\textwidth]{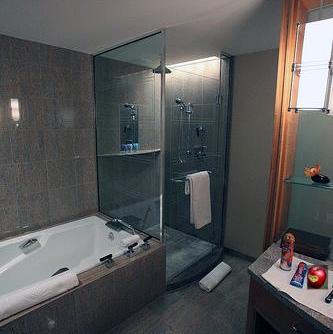}
        \caption{a bathroom with a bathtub}
    \end{subfigure}
    \caption{Examples where the VQA model predicted the incorrect answer. The VQA model often responds with an ordinary object in the scene, apparently not understanding the point of the question.}
    \label{fig:example_response}
\end{figure}
\begin{figure}
\begin{minipage}[!t]{\textwidth}
  \begin{minipage}[!t]{0.49\textwidth}
    \centering
    \includegraphics[width=.95\textwidth]{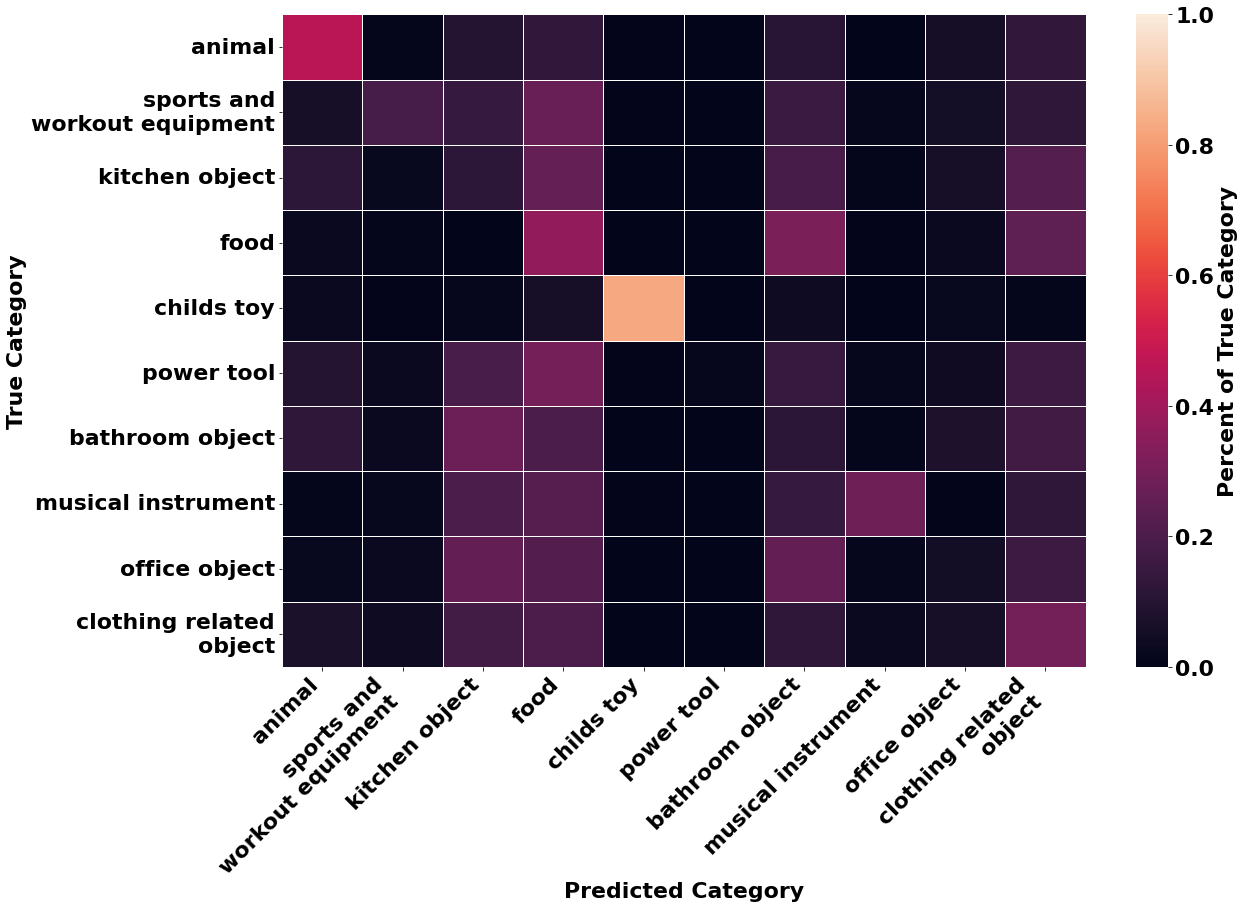}
    \captionof{figure}{Confusion matrix for the VQA broad category matching metric.}
    \label{fig:broad_confusion}
  \end{minipage}
  \hfill
  \begin{minipage}[b]{0.49\textwidth}
    \centering
    \captionof{table}{Performance of similarity-based method on contextual anomaly detection}
    \begin{tabular}{lccl}\toprule 
    & \multicolumn{2}{c}{Accuracy}
    \\\cmidrule(lr){2-3}
    \makecell{Set of Similarity \\ Functions}  & Top 1  & Top 3 \\\midrule
    All       & 38.00 & 38.61  \\
    Visual    & 36.31 & 36.91 \\
    Knowledge & 12.24 & 12.63 \\
     \bottomrule
    \end{tabular}
    \label{tab:sim_acc}
  \end{minipage}
\end{minipage}
\end{figure}

\subsection{VQA-based method}

Table \ref{tab:blip_acc} shows the performance of each variant of the BLIP-2 model using each evaluation method. However, on all evaluation methods the VQA models performed significantly poorly. The XXL model was the best performing model, but only achieved achieved  12.32\% accuracy on the VQA-like metric and 13.41\% performance on the matching metric. Additionally, a qualitative review of incorrect responses indicates that the model often selected a random, non-anomalous object as the anomaly. Examples of model responses are shown in Fig.  \ref{fig:example_response}. This behavior aligns with the broad category matching results; if the model was responding with semantically correct answers, the accuracy on the broad category matching metric would be high. Instead, the best VQA model only achieved 22.13\% top 1 accuracy. Given that there are only 10 possible categories for this metric, the VQA model only marginally outperforms a random baseline. The broad category confusion matrix, given in Figure \ref{fig:broad_confusion}, further highlights this behavior; the model only beats 50\% performance on the animal and child's toy categories, consistently predicting the wrong category for all other classes.



\subsection{Similarity-based anomaly detection}

Table \ref{tab:sim_acc} records the full results for the similarity based anomaly detection method; note that this metric corresponds to the VQA matching metric as both metrics make predictions over the Open Images objects. The similarity-based anomaly detection method achieved a maximum top 1 accuracy of 38.00\% on the generated anomaly dataset. Accuracy was highest when using all similarity functions, dropping slightly to 36.31\% when only visual similarity functions were used and dropping significantly to 12.24\% when only the knowledge based similarity function was used. 

\section{Discussion and conclusions}

Since the data generation pipeline was computationally lightweight, the resulting dataset of context-dependent anomalies was still much larger than any comparable dataset, even though a substantial fraction of generated samples did not pass the content filter. Qualitatively, most samples that passed the content filter during evaluation align with the intended label, and were smoothly integrated into the image. However, artifacts are still present in the accepted images; in particular, the pose or size of the inserted object is often slightly wrong, and in some cases the background of the inserted visual information does not match surrounding image. Future work should consider improved methods for filtering out these artifacts; however, while undesirable, these artifacts do not change the apparent class of the injected visual data. This ensures that data containing these artifacts remains anomalous, and thus does not prevent the generated dataset from serving as an anomaly detection benchmark. In all likelihood artifacts should make the detection of anomalies easier, as they introduce other information which can potentially be used to detect the anomalous region.

Context-dependent anomaly detection is a challenging task; however, the similarity-based anomaly detection method was able to identify a reasonable fraction of anomalies despite using a rudimentary method to reason about the anomalies. The method performed significantly worse when relying on only semantic features as opposed to visual features, and further experimentation is needed to gain more insight into why semantic features underperformed. However, the overall performance of the similarity based method indicated that the quality of the synthetic data sufficient for the anomaly detection task to be tractable. 


All variants of the BLIP-2 VQA model test performed poorly on the synthetically generated anomaly dataset, regardless of evaluation technique. It is unlikely that this failure stemed from BLIP-2's vision module, as the image encoder it uses is similar to the models used by the similarity-based method. Since the synthetic data is qualitatively convincing and is likely tractable, it is much more likely that the reasoning capabilities of the underlying LLM used by BLIP were not aligned with the context-dependent anomaly detection task we posed. This result contrasts with BLIP-2s competitive zero-shot performance on the VQAv2 dataset (65\%) and the Open Knowledge VQA dataset (45.9\%), and indicates that these datasets may not provide challenging benchmarks for tasks which require reasoning about anomalies.

The ability of the synthetic data to challenge the VQA model on the context-dependent anomaly detection task provides a highlights the potential for synthetic data to provide insight on other challenging reasoning tasks. For example, with minimal adjustments our pipeline could be used to generate data for object state queries (e.g. has the the apple been eaten) or task completion queries (e.g. has the bed been made). Our framework could also be used for augmenting existing data for class balance (i.e. replacing a common dog breed with a rare one) or robustness (e.g. changing the type of utensils at a table for a semantic distribution shift), but additional work is needed to determine whether our method is a viable approach for these augmentation tasks. The primary limitation of our method is the capability of the the image generation model used; if the model cannot generate high quality instances of the target object in a specified state, the quality of the generated dataset may be insufficient to serve as a useful benchmark for the task. An appropriate method for filtering out undesirable samples must also be available; while the process we defined in this work is applicable to data generation for a variety of tasks, there may be tasks for which a different filtering method must be developed.


Our work demonstrates that it is possible to generate a challenging, task specific bench mark for a multi-modal model using existing text-to-image generation capabilities at a lower cost. We do not expect this capability to replace the traditional paradigm of dataset development that has powered the development of deep learning models, as naturally occurring data is by definition the gold standard for evaluation. However, our pipeline offers a low-cost diagnostic tool that can be used to better understand the current capability of AI models on tasks, like context-dependent anomaly detection, for which datasets of naturally-occuring data do not exist. This may allow researchers to gain more insight into the performance gaps of current models and ultimately develop new architectures and training methods to address these gaps. 





\begin{ack}
DISTRIBUTION STATEMENT A. Approved for public release. Distribution is unlimited.

This material is based upon work supported by the Department of the Air Force under Air Force Contract No. FA8702-15-D-0001. Any opinions, findings, conclusions or recommendations expressed in this material are those of the author(s) and do not necessarily reflect the views of the Department of the Air Force.

© 2023 Massachusetts Institute of Technology.

Delivered to the U.S. Government with Unlimited Rights, as defined in DFARS Part 252.227-7013 or 7014 (Feb 2014). Notwithstanding any copyright notice, U.S. Government rights in this work are defined by DFARS 252.227-7013 or DFARS 252.227-7014 as detailed above. Use of this work other than as specifically authorized by the U.S. Government may violate any copyrights that exist in this work.

The authors would like to thank Dr. Sanjeev Mohindra and Ms. Paula Donovan for their support and Drs. Rajmonda Caceres and Kevin Leahy for their time and helpful feedback. 
\end{ack}

\small{
 \bibliographystyle{abbrvnat}
\bibliography{main}
}

\newpage
\appendix

\section{Additional examples of generated synthetic data}
\begin{figure}[!h]
    \centering
    \begin{subfigure}[b]{1.00\textwidth}
        \centering
        \includegraphics[width=.24\textwidth]{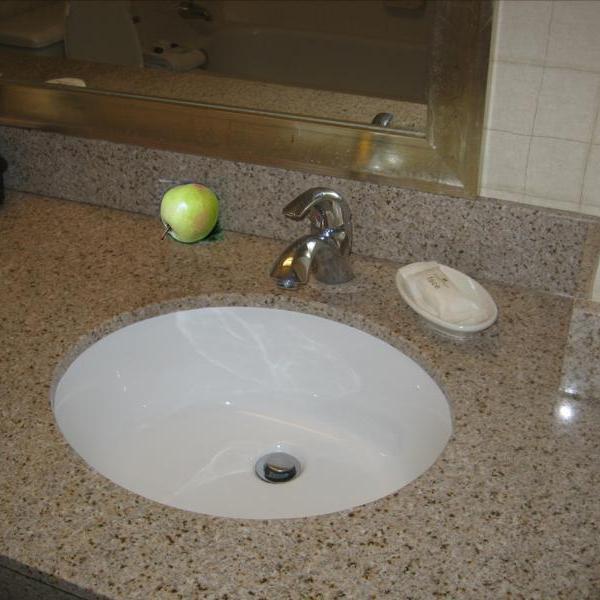}
        \includegraphics[width=.24\textwidth]{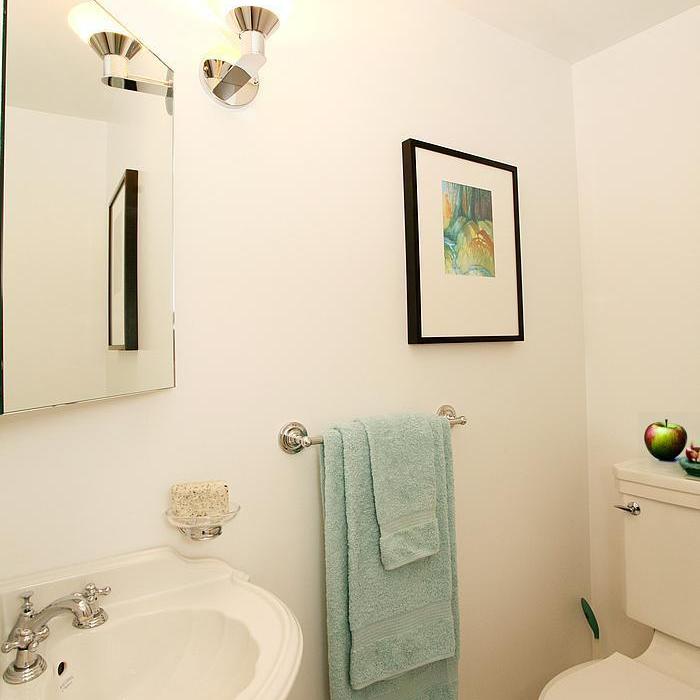}
        \includegraphics[width=.24\textwidth]{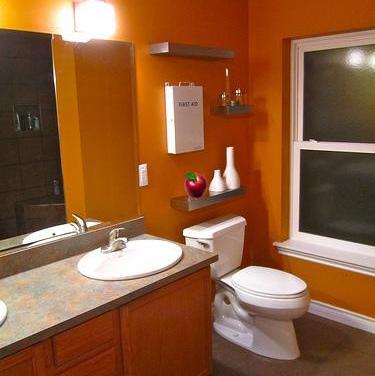}
        \includegraphics[width=.24\textwidth]{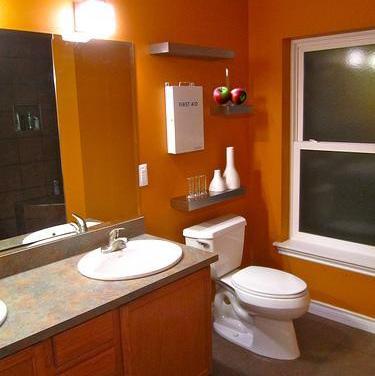}
        \caption{bathrooms with anomalies}
    \end{subfigure}
    \\[\smallskipamount]
    \begin{subfigure}[b]{1.00\textwidth}
        \centering
        \includegraphics[width=.24\textwidth]{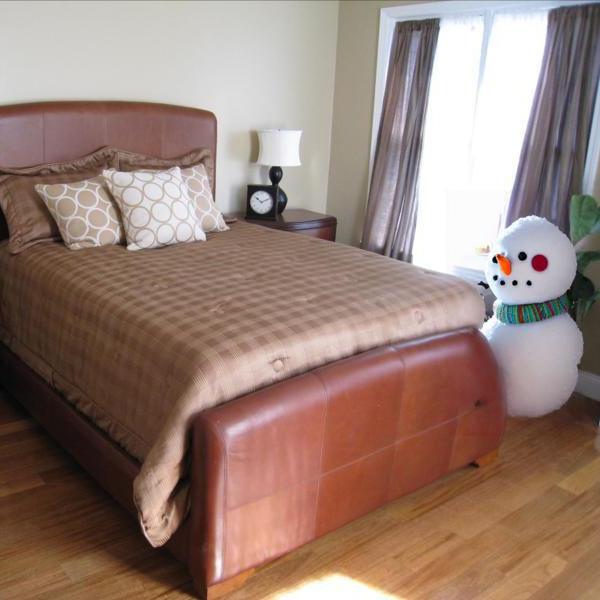}
        \includegraphics[width=.24\textwidth]{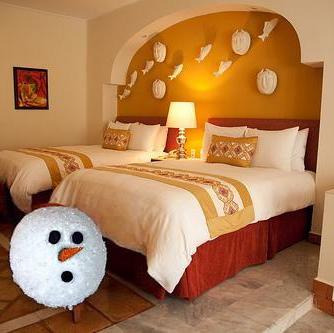}
        \includegraphics[width=.24\textwidth]{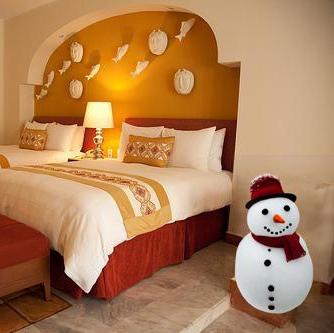}
        \includegraphics[width=.24\textwidth]{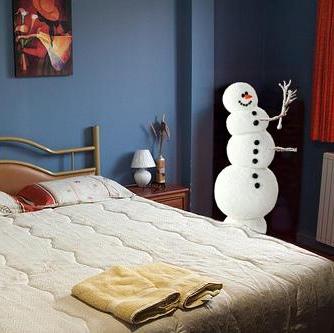}
        \caption{hotel rooms with anomalies}
    \end{subfigure}
    \\[\smallskipamount]
    \begin{subfigure}[b]{1.00\textwidth}
        \centering
        \includegraphics[width=.24\textwidth]{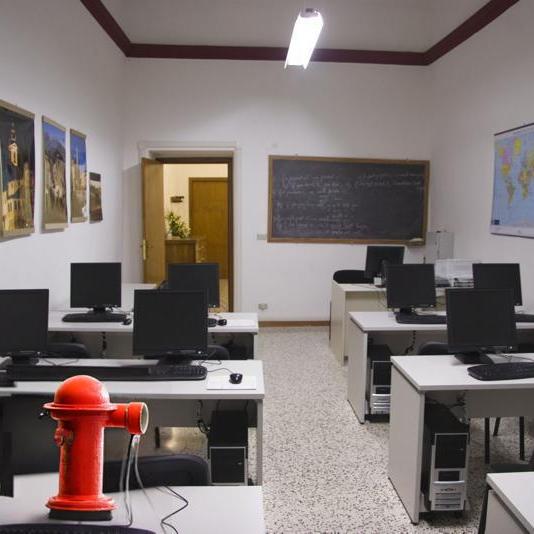}
        \includegraphics[width=.24\textwidth]{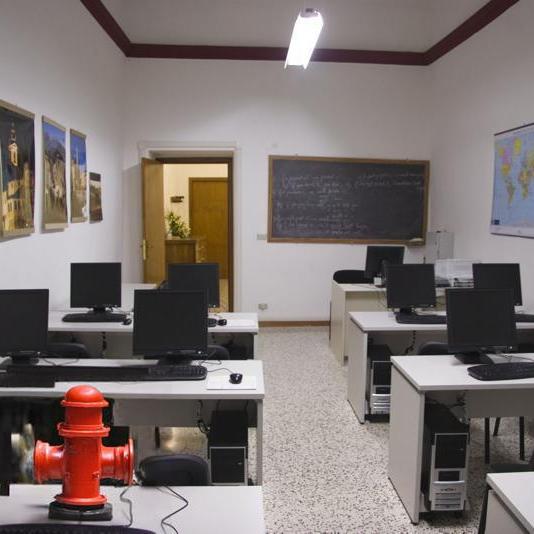}
        \includegraphics[width=.24\textwidth]{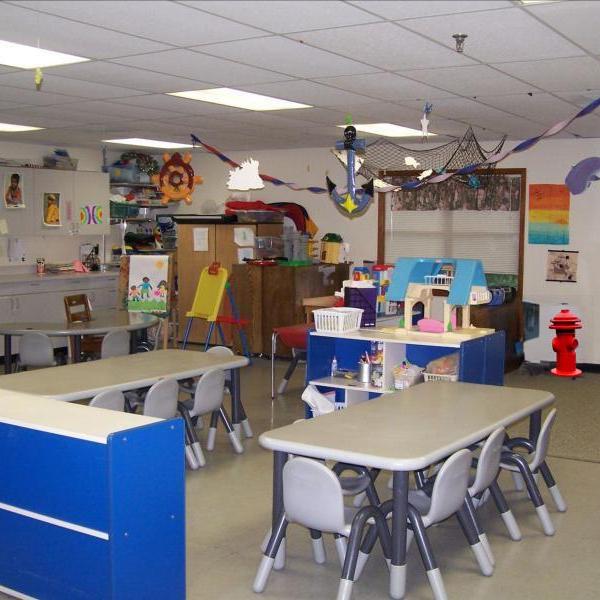}
        \includegraphics[width=.24\textwidth]{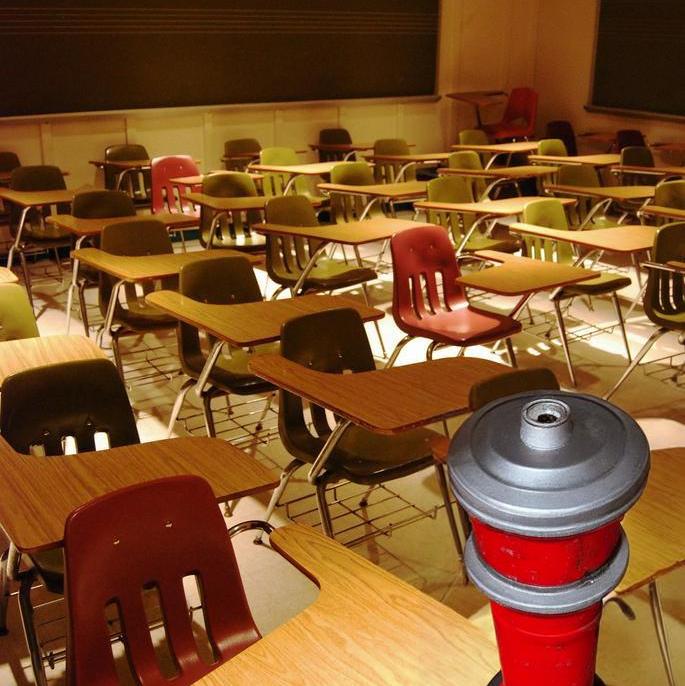}
        \caption{classrooms with anomalies}
    \end{subfigure}
    \\[\smallskipamount]
    \begin{subfigure}[b]{1.00\textwidth}
        \centering
        \includegraphics[width=.24\textwidth]{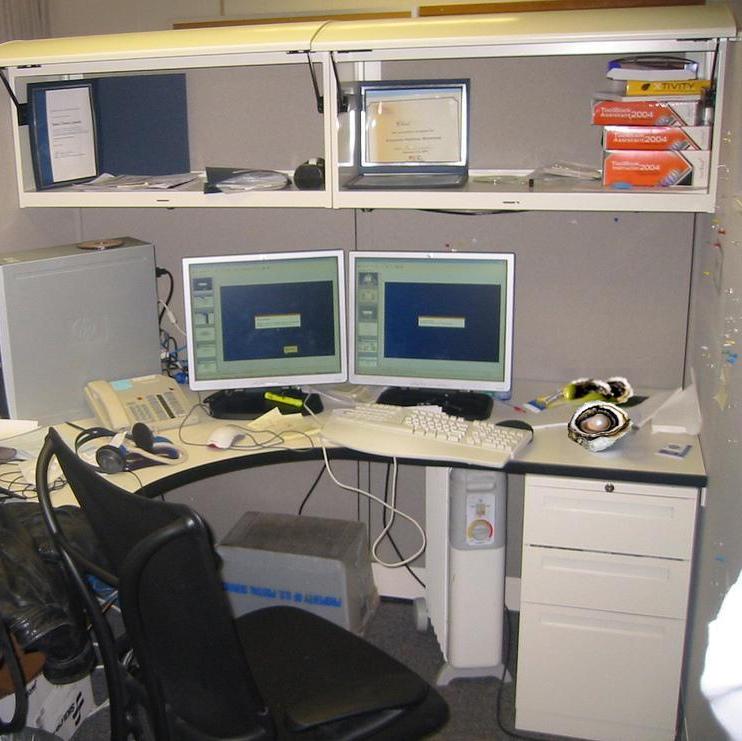}
        \includegraphics[width=.24\textwidth]{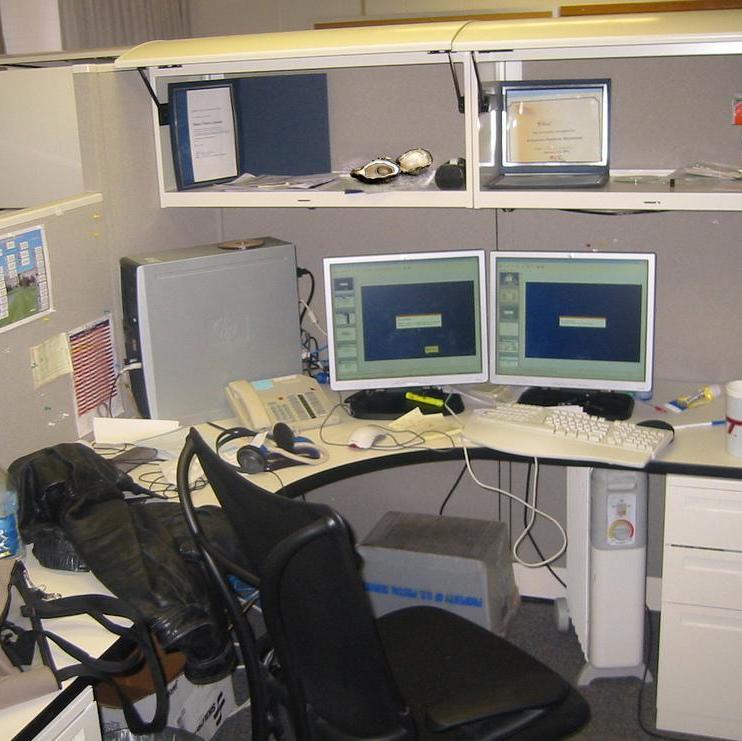}
        \includegraphics[width=.24\textwidth]{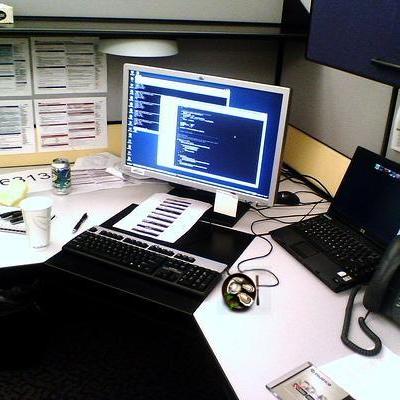}
        \includegraphics[width=.24\textwidth]{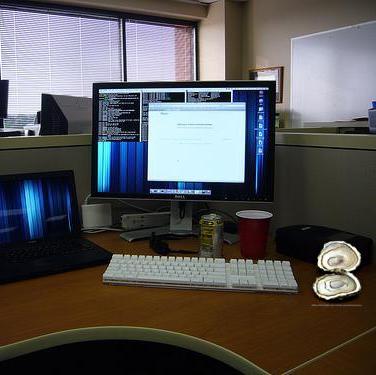}
        \caption{cubicles with anomalies}
    \end{subfigure}
    \caption{Additional examples of generated anomalous data for selected scene types}
    \label{fig:add_gen1}
    \end{figure}

    \begin{figure}[!h]
    \centering
        \begin{subfigure}[b]{1.00\textwidth}
        \centering
        \includegraphics[width=.24\textwidth]{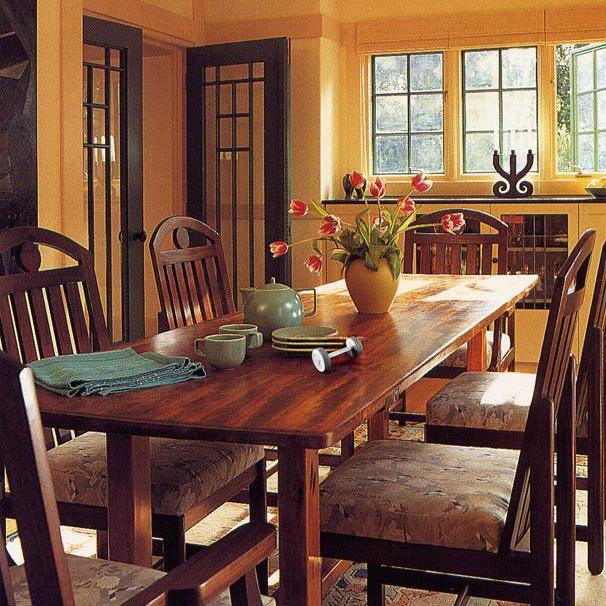}
        \includegraphics[width=.24\textwidth]{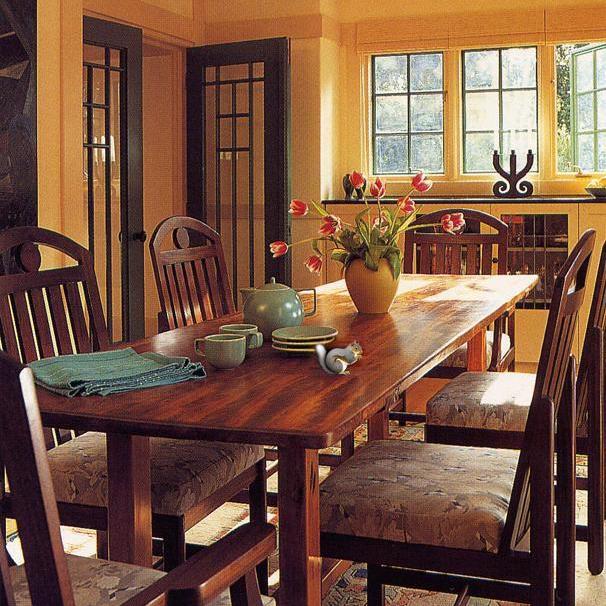}
        \includegraphics[width=.24\textwidth]{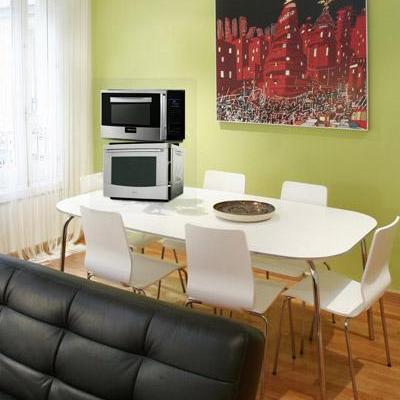}
        \includegraphics[width=.24\textwidth]{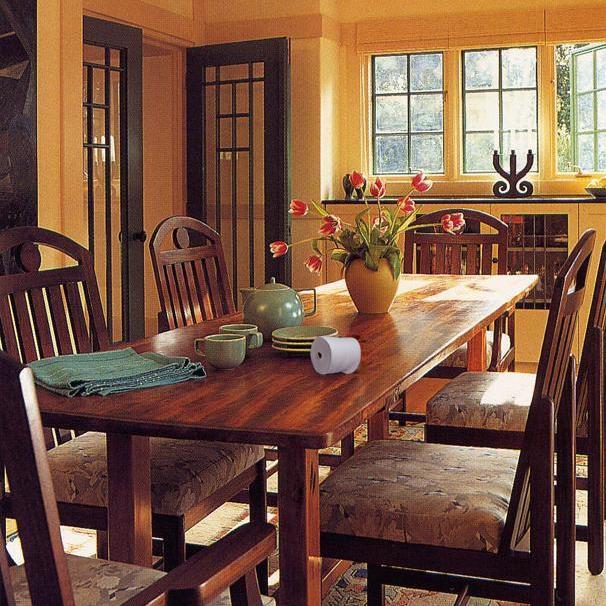}
        \caption{dining rooms with anomalies}
    \end{subfigure}
    \\[\smallskipamount]
        \begin{subfigure}[b]{1.00\textwidth}
        \centering
        \includegraphics[width=.24\textwidth]{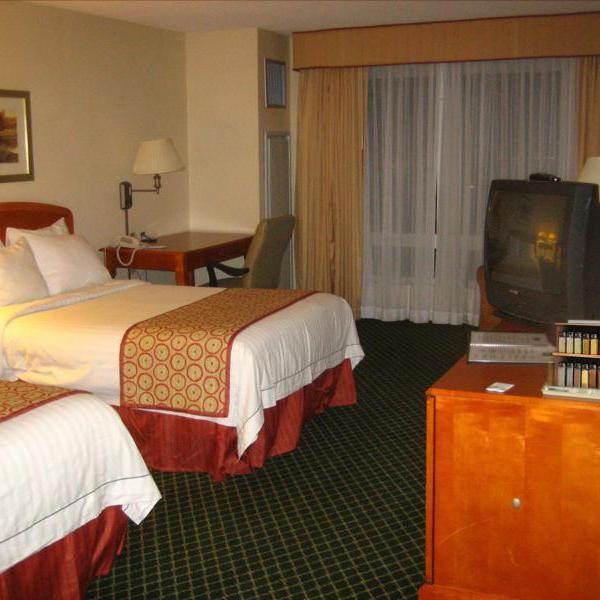}
        \includegraphics[width=.24\textwidth]{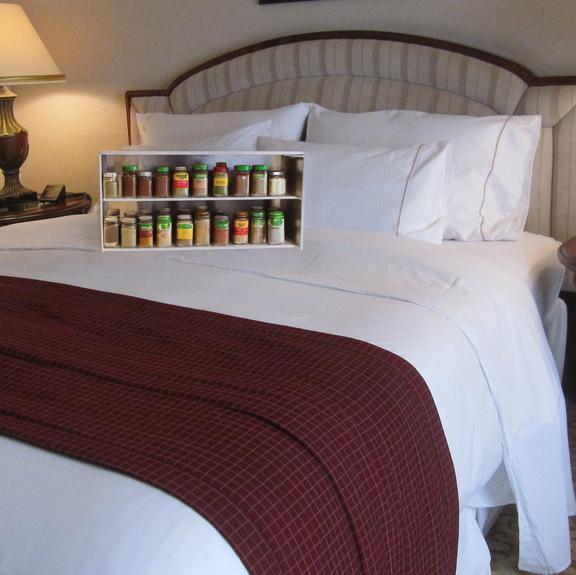}
        \includegraphics[width=.24\textwidth]{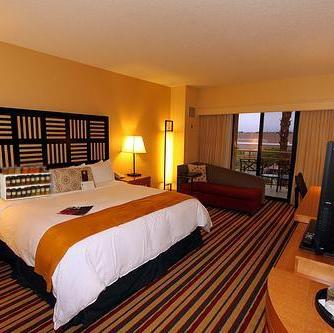}
        \includegraphics[width=.24\textwidth]{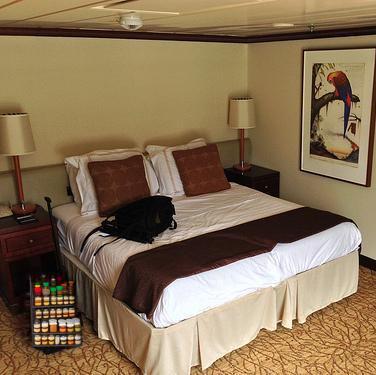}
        \caption{hotel rooms with anomalies}
    \end{subfigure}
    \\[\smallskipamount]
        \begin{subfigure}[b]{1.00\textwidth}
        \centering
        \includegraphics[width=.24\textwidth]{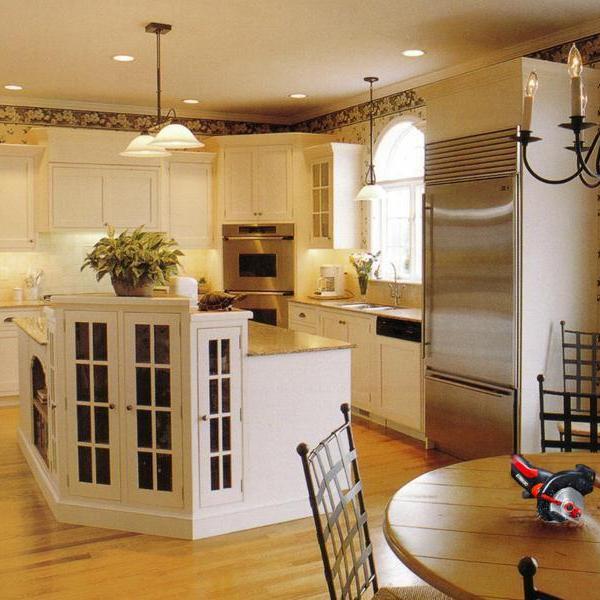}
        \includegraphics[width=.24\textwidth]{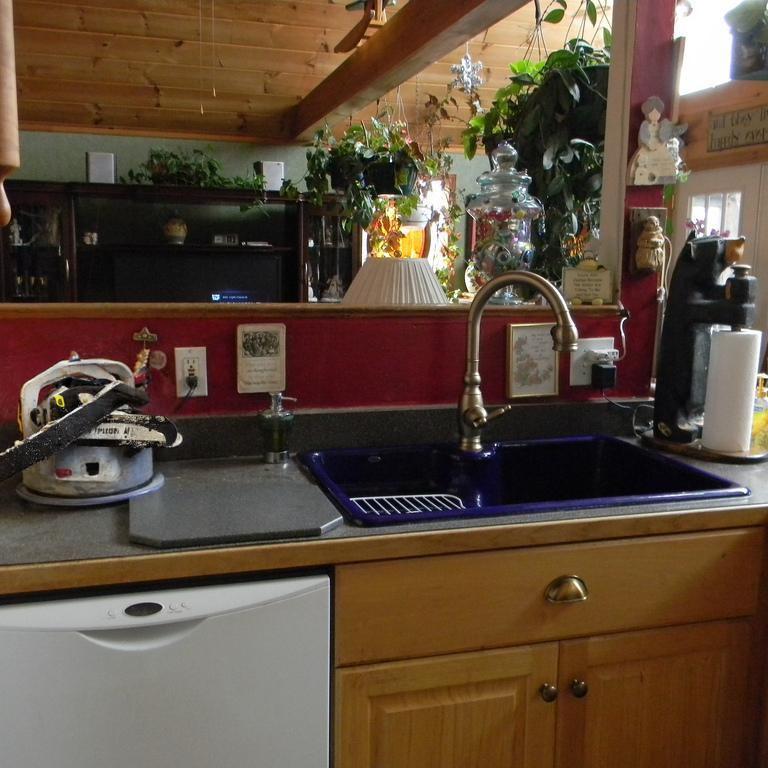}
        \includegraphics[width=.24\textwidth]{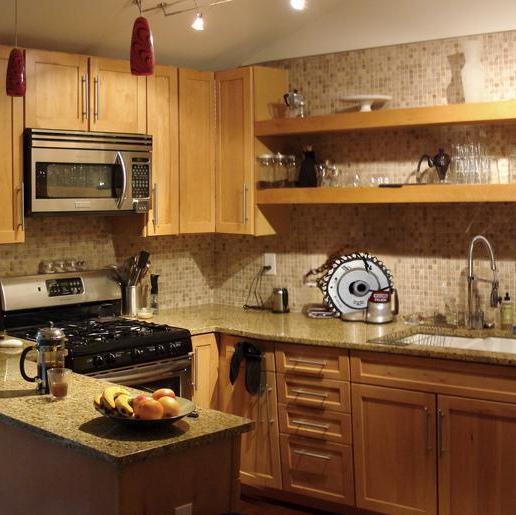}
        \includegraphics[width=.24\textwidth]{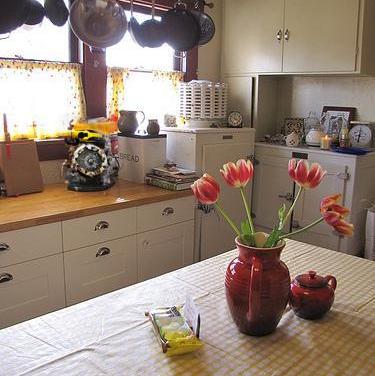}
        \caption{kitchens with anomalies}
    \end{subfigure}
    \\[\smallskipamount]
        \begin{subfigure}[b]{1.00\textwidth}
        \centering
        \includegraphics[width=.24\textwidth]{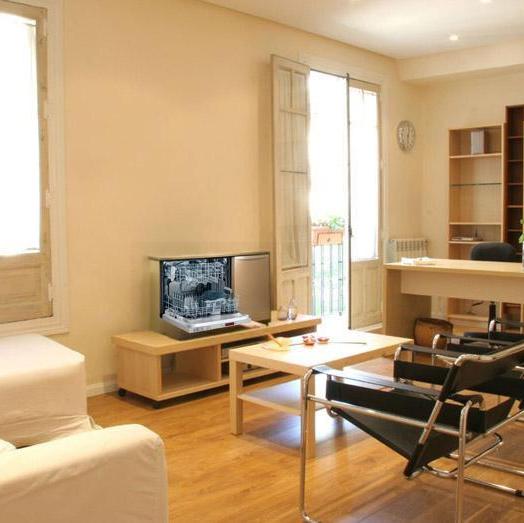}
        \includegraphics[width=.24\textwidth]{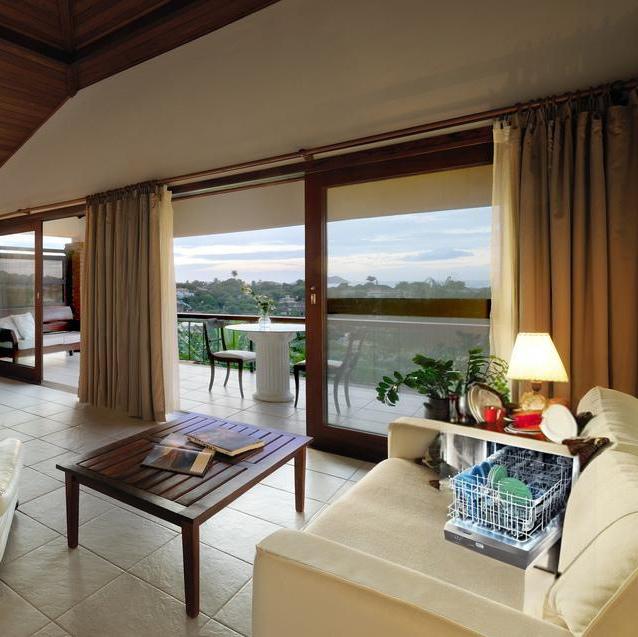}
        \includegraphics[width=.24\textwidth]{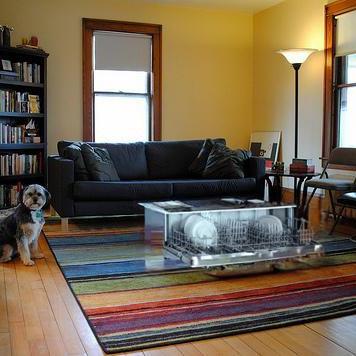}
        \includegraphics[width=.24\textwidth]{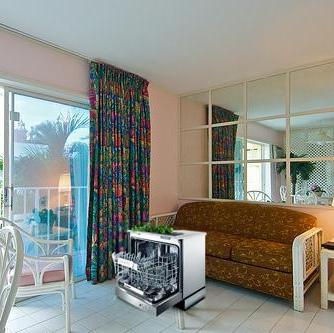}
        \caption{living rooms with anomalies}
    \end{subfigure}
    \caption{Additional examples of generated anomalous data for remaining scene types}
    \label{fig:add_gen2}
\end{figure}

Figures \ref{fig:add_gen1} and \ref{fig:add_gen2} show additional examples of anomalies generated using our synethetic data generation framework. Although relatively few underlying natural images are used, diversity is introduced through the choice of mask, anomaly, and text-to-image generator random seed, resulting in a substantial amount of variation in the generated images. 

\begin{figure}[!h]
    \centering
    \begin{subfigure}[b]{0.24\textwidth}
        \centering
        \includegraphics[width=.9\textwidth]{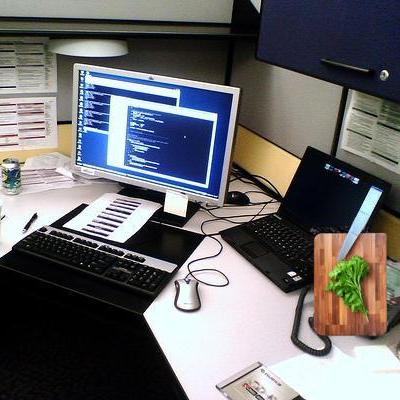}
        \caption{\makecell[l]{Incorrect object pose  \\ with respect to scene}}
    \end{subfigure}
    \begin{subfigure}[b]{0.24\textwidth}
        \centering
        \includegraphics[width=.9\textwidth]{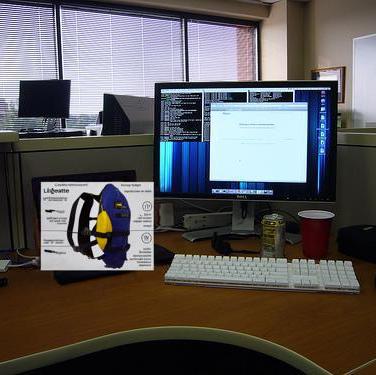}
        \caption{\makecell[l]{Incorrect background\\ infilling}}
    \end{subfigure}
    \begin{subfigure}[b]{0.24\textwidth}
        \centering
        \includegraphics[width=.9\textwidth]{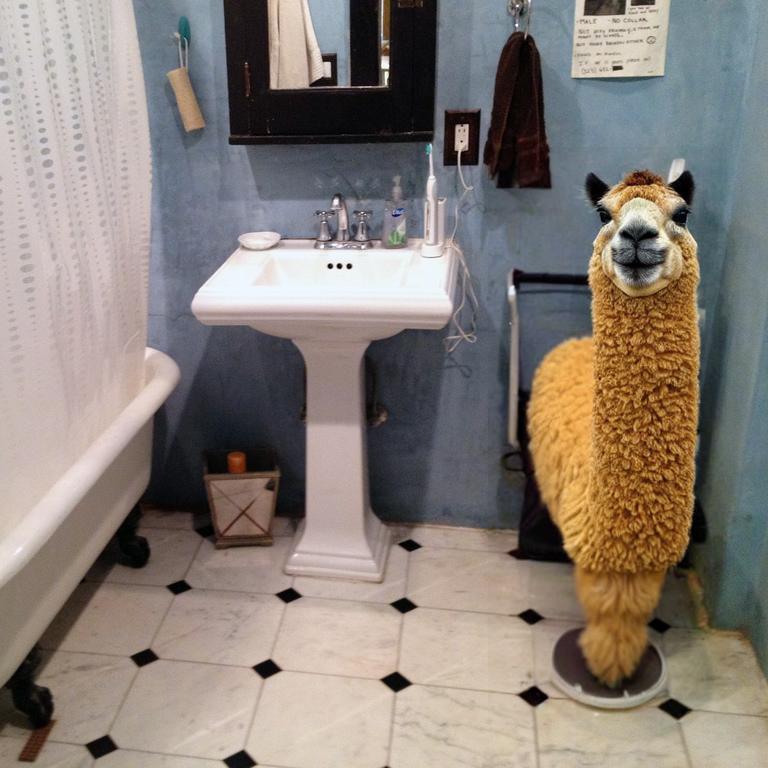}
        \caption{\makecell[l]{Significant distortion \\in object}}
    \end{subfigure}
    \begin{subfigure}[b]{0.24\textwidth}
        \centering
        \includegraphics[width=.9\textwidth]{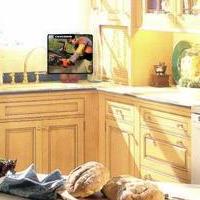}
        \caption{\makecell[l]{Human was not\\ filtered out}}
    \end{subfigure}
    \caption{Common artifact types in generated images}
    \label{fig:artifacts}
\end{figure}

Due to the limitations of current text-to-image generation models, some of the generated images contain artifacts. Figure \ref{fig:artifacts} shows common types of artifacts; these include incorrectly posed objects, objects in which the background does not match the surrounding scene, objects which are significantly distorted, and generations which include a human or parts of humans (i.e., limbs). While these artifacts are undesirable, they also make the anomaly detection problem easier by introducing point anomaly features which could be provide additional information during evaluation. However, as the performance of the VQA models is still poor, the generated data appears to still be challenging despite the occasional presence of these artifacts.

\begin{figure}[!h]
    \centering
    \begin{subfigure}[b]{1.00\textwidth}
        \centering
        \includegraphics[width=.24\textwidth]{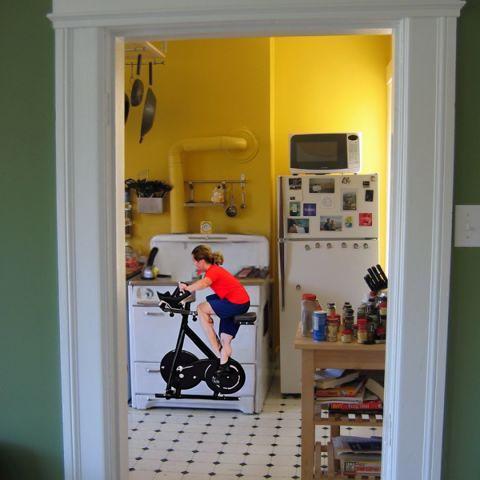}
        \includegraphics[width=.24\textwidth]{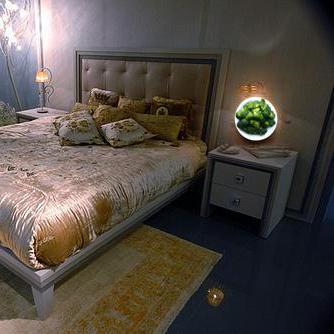}
        \includegraphics[width=.24\textwidth]{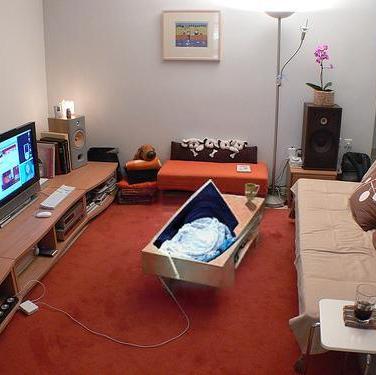}
        \includegraphics[width=.24\textwidth]{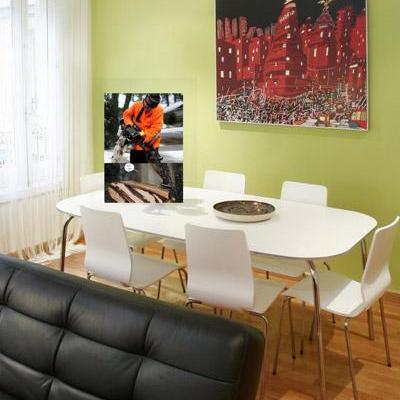}
        \caption{Incorrect generations which were correctly filtered out}
    \end{subfigure}
    \\[\smallskipamount]
    \begin{subfigure}[b]{1.00\textwidth}
        \centering
        \includegraphics[width=.24\textwidth]{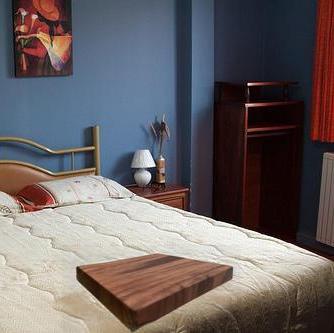}
        \includegraphics[width=.24\textwidth]{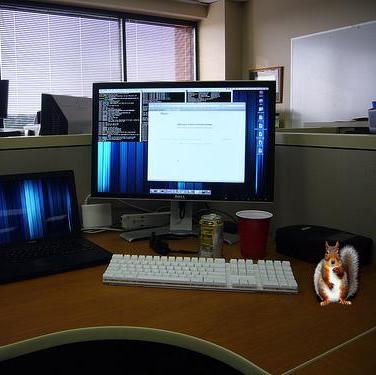}
        \includegraphics[width=.24\textwidth]{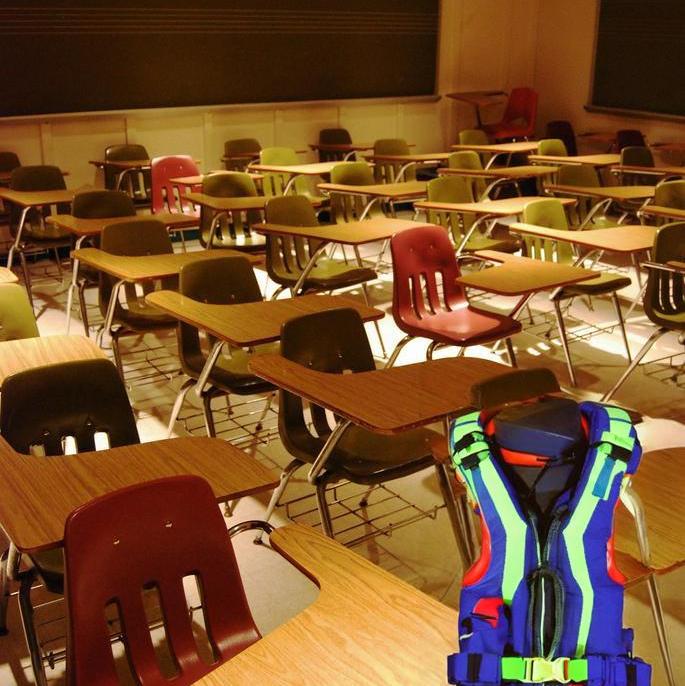}
        \includegraphics[width=.24\textwidth]{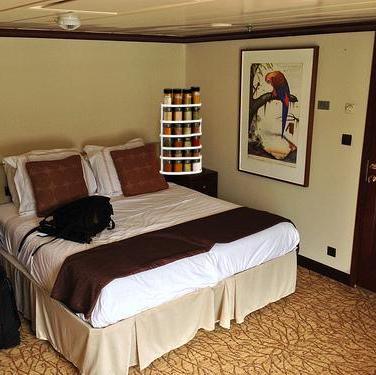}
        \caption{Correct generations which were incorrectly filtered out}
    \end{subfigure}
    \caption{Example generated images that were rejected by content filters}
    \label{fig:rejects}
\end{figure}

Figure \ref{fig:rejects} shows example generations that were rejected by the content filters. In many cases, filter decisions are reasonable; removing images with humans in them or ones in which the target object (e.g.: cutting board, washing machine) was not accurately produced. However, the filters also remove instances in which the generation is reasonable. This does not pose a significant issue for our framework; since data generation is computationally lightweight, the images removed by conservative filters can be easily replaced without.

\newpage

\section{Objects, anomaly definitions, and scene types}

Table \ref{tab:object_anom} shows the full list of objects and scenes. Scenes types were selected based by their availability in the Visual Genome dataset. Objects were selected from the set of objects included in the Open Images object dataset to provide coverage of different object sizes and types. Objects were also selected to ensure that each scene type had some anomalous object types associate with it; this classification was done manually to ensure consistency in anomaly definition. There are also a few objects included in the dataset like llama or traffic light which are considered to be anomalous in every scene type; these were included to expand the space of anomalies considered. 

\begin{table}[!h]
\caption{Object anomaly  }
\centering
\begin{tabular}{lcccccccccl}\toprule 

& \multicolumn{8}{c}{Scene Type} & \\\cmidrule(lr){2-9}

Object        & bathroom  & bedroom     & classroom    & cubicle  & \makecell{dining \\room} & \makecell{hotel \\room} & kitchen & \makecell{living \\room} \\\midrule
apple &  X & X &   &   &   &   &   &   \\
alpaca &  X & X & X & X & X & X & X & X \\
\makecell[l]{baseball \\ glove} &  X &   &   & X & X &   & X    \\
binoculars &  X & X &   & X & X & X & X &   \\
bidet &    & X & X & X & X & X & X & X \\
chainsaw &  X & X & X & X & X & X & X & X \\
\makecell[l]{computer} &  X &   &   &   & X & X & X &   \\
crocodile &  X & X & X & X & X & X & X & X \\
\makecell[l]{cutting board} &  X & X & X & X &   & X &   & X \\
dishwasher &  X & X & X & X & X & X &   & X \\
doll &  X &   &   & X & X & X & X &   \\
drum &  X & X &   & X & X & X & X &   \\
dumbbell &  X &   & X & X & X & X & X &   \\
\makecell[l]{fire hydrant} &  X & X & X & X & X & X & X & X \\
fish &  X & X & X & X &   & X &   &   \\
\makecell[l]{helmet} &  X & X & X & X & X & X & X &   \\
\makecell[l]{frying pan} &  X & X & X & X &   & X &   & X \\
golf ball &  X &   &   & X & X & X & X &   \\
hamburger &  X & X &   &   &   &   &   &   \\
harpsichord &  X & X &   & X & X & X & X &   \\
lifejacket &  X & X & X & X & X & X & X & X \\
\makecell[l]{microwave} &  X & X &   &   & X & X &   & X \\
\makecell[l]{office \\ supplies} &  X &   &   &   &   &   & X &   \\
oyster &  X & X & X & X &   & X &   & X \\
panda &  X & X & X & X & X & X & X & X \\
perfume &    &   & X &   & X &   & X & X \\
pizza &  X & X &   &   &   &   &   &   \\
printer &  X & X &   &   & X &   & X &   \\
\makecell[l]{punching \\ bag} &  X & X & X & X & X & X & X &   \\
rocket &  X &   &   & X & X & X & X &   \\
saxophone &  X &   &   & X & X & X & X &   \\
\makecell[l]{sewing \\ machine} &  X &   &   & X & X & X & X &   \\
slow cooker &  X & X & X & X &   & X &   & X \\
snowman &  X & X & X & X & X & X & X & X \\
spice rack &  X & X & X & X & X & X &   & X \\
spoon &  X & X &   &   &   &   &   &   \\
squirrel &  X & X & X & X & X & X & X & X \\
\makecell[l]{stationary \\ bicycle} &  X & X & X & X & X & X & X &   \\
stethoscope &  X & X & X & X & X & X & X & X \\
stretcher &  X & X & X & X & X & X & X & X \\
surfboard &  X & X & X & X & X & X & X & X \\
syringe &  X & X & X & X & X & X & X & X \\
toilet paper &    & X & X & X & X & X & X & X \\
toothbrush &    & X & X & X & X & X & X & X \\
towel &    & X & X & X & X &   &   & X \\
traffic light &  X & X & X & X & X & X & X & X \\
violin &  X &   &   & X & X & X & X &   \\
\makecell[l]{washing \\ machine} &  X & X & X & X & X & X & X & X \\
\cmidrule(lr){1-9}
 
\end{tabular}
\label{tab:object_anom}
\end{table}

\newpage
\section{Additional details on method}

Text descriptions of objects were needed for multiple components of our methods, including the similarity function knowledge base, the object descriptions for improving matching accuracy, and the answer descriptions for  matching. We used the same text completion prompting regardless of the the text generation model used. This prompt involved several human-written descriptions for other objects (e.g. "dog: a type of pet...") followed by the target text for which a description was needed and a blank space. We found that this prompting consistently generated high quality descriptions, but in order to ensure that the knowledge base was accurate, all knowledge-based entries were read and regenerated if any factual or logical errors were noticed.  

 Three different similarity functions were defined to compute similarity scores for each region; image-region visual similarity ($IRV$), region-region visual similarity ($RRV$), and knowledge-based similarity ($KB$). All three similarity functions utilized a joint visual-language embedding  model $f$ to compute similarity scores.
 
 Let $f(I) = e_I$ be the image embedding (i.e., the visual features of the entire image). Let $f(r) = e_r$ be the visual embedding for a region, and $E_r$ be the set of visual embeddings of all regions. 
 
 The $IRV$ similarity function calculates embeddings for the visual features of the full image and each region; the more similar the region's embedding is to the full image embedding, the higher the similarity score: 
 
 $$ IRV(e_r) = e_i \cdot e_r, \forall e_r \in E_r$$

The $RRV$ similarity function only compares visual embeddings between regions, and calculates a similarity score based on how similar an individual regions features are from all other regions:

$$ RRV(e_r) = \frac{1}{|E_r|}\sum_{e \in E_r} e_r \cdot e  $$

The $KB$ similarity function compares the semantic information associated with a region against the semantic information associated with other regions. The score represents how semantically close the knowledge associated with each region is to the overall semantic context of the scene. Let $f(o) = e_o$ be the semantic embedding generated from the natural language description of an object class, and $E_o$ be the set of semantic embeddings for all object classes. We first calculate the similarity $s_o$ between the visual features for the region and the semantic embedding of an object class, repeating for all object classes:

$$ s_o = e_r \cdot e_o, \forall e_o \in E_o  $$

Object descriptions corresponding to the top-$k$ values of all $s_o$ (i.e., most similar classes to each region) are retrieved from a knowledge base and concatenated into a text we denote as $d$; an additional set of embeddings for these joint descriptions is calculated as $f(d) = e_d$ for each region. The set of these description embeddings is written as $E_d$, and a region-to-region similarity score is calculated with these embeddings:

$$ KB(e_d) =\frac{1}{|E_d|}\sum_{e \in E_d} e_d \cdot e $$

\end{document}